\documentclass{sig-alternate-2013}

\usepackage{subfigure, float, graphicx, epstopdf, balance}
\usepackage{color, soul}
\usepackage[table]{xcolor}
\usepackage{epsfig}
\usepackage{booktabs}

\usepackage{mathtools, url}
\usepackage{latexsym}
\usepackage{amsfonts} 
\usepackage{adjustbox}
\usepackage{lipsum}

\def\P{\mathbb{P}}

\usepackage{float} 
\usepackage{amsmath}
\usepackage{amssymb}
\usepackage{amsfonts}

\usepackage{multirow}

\usepackage[table]{xcolor}
\usepackage{epsfig}
\usepackage{booktabs, balance}
\usepackage{adjustbox, url}

\def\P{\mathbb{P}}

\definecolor{lightgray}{gray}{0.85}

\definecolor{lightgray}{gray}{0.85}

\permission{Copyright is held by the International World Wide Web Conference Committee (IW3C2). IW3C2 reserves the right to provide a hyperlink to the author's site if the Material is used in electronic media.}
\conferenceinfo{WWW 2015,}{May 18--22, 2015, Florence, Italy.}
\copyrightetc{ACM \the\acmcopyr}
\crdata{978-1-4503-3469-3/15/05. \\
http://dx.doi.org/10.1145/2736277.2741643}

\begin{document}

\title{Hierarchical Neural Language Models for Joint Representation of Streaming Documents and their Content}

\author{Nemanja Djuric\thanks{Authors contributed equally. This work was done while Hao Wu was an intern at Yahoo Labs in Sunnyvale, USA.} $^\dag$, Hao Wu$^{*\ddag}$, Vladan Radosavljevic$^\dag$, Mihajlo Grbovic$^\dag$, Narayan Bhamidipati$^\dag$\\
\affaddr{$^\dag$Yahoo Labs, Sunnyvale, CA, USA, \{nemanja, vladan, mihajlo, narayanb\}@yahoo-inc.com} \\
\affaddr{$^\ddag$University of Southern California, Los Angeles, CA, USA, hwu732@usc.edu} \\
}



\maketitle
\begin{abstract}
We consider the problem of learning distributed representations for documents in data streams. The documents are represented as low-dimensional vectors and are jointly learned with distributed vector representations of word tokens using a hierarchical framework with two embedded neural language models. In particular, we exploit the context of documents in streams and use one of the language models to model the document sequences, and the other to model word sequences within them. The models learn continuous vector representations for both word tokens and documents such that semantically similar documents and words are close in a common vector space. We discuss extensions to our model, which can be applied to personalized recommendation and social relationship mining by adding further user layers to the hierarchy, thus learning user-specific vectors to represent individual preferences. We validated the learned representations on a public movie rating data set from MovieLens, as well as on a large-scale Yahoo News data comprising three months of user activity logs collected on Yahoo servers. The results indicate that the proposed model can learn useful representations of both documents and word tokens, outperforming the current state-of-the-art by a large margin.
\end{abstract}

\category{I.2.7}{Artificial Intelligence}{Natural Language Processing}[Language Models]
\category{I.5.4}{Pattern Recognition}{Applications}[Text processing]
\category{I.7.m}{Document and Text Processing}{Miscellaneous}


\keywords{Machine learning; document modeling; distributed representations; word embeddings; document embeddings.} 

\section{Introduction}
Text documents coming in a sequence are common in real-world applications and can arise in various contexts. For example, consider Web pages surfed by users in random walks along the hyperlinks, streams of click-through URLs associated with a query in search engine, publications of an author in chronological order, threaded posts in online discussion forums, answers to a question in online knowledge-sharing communities, or e-mails in a common thread, to name a few. The co-occurrences of documents in a temporal sequence may reveal relatedness between them, such as their semantic and topical similarity. In addition, sequences of words within the documents introduce another rich and complex source of the data, which can be leveraged together with the document stream information to learn useful and insightful representations of both documents and words.

In this paper, we introduce algorithm that can simultaneously model documents from a stream and their residing natural language in one common lower-dimensional vector space. Such algorithm goes beyond representations which consider each document as a separate bag-of-words composition, most notably Probabilistic Latent Semantic Analysis (PLSA) \cite{hofmann1999probabilistic} and Latent Dirichlet Allocation (LDA) \cite{blei2003latent}. We focus on learning continuous representations of documents in vector space jointly with distributed word representations using statistical neural language models \cite{bengio2003neural}, whose success was previously shown in a number of publications \cite{collobert2008unified}. More precisely, we propose hierarchical models where document vectors act as units in a context of document sequences, and also as global contexts of word sequences contained within them. The probability distribution of observing a word depends not only on some fixed number of surrounding words, but is also conditioned on the specific document. Meanwhile, the probability distribution of a document depends on the surrounding documents in stream data. 
Our work is most similar to \cite{le2014distributed} as it also considers document vectors as a global context for words contained within, however the authors in \cite{le2014distributed} do not model the document relationships which brings significant modeling strength to our models and opens a plethora of application areas for the neural language models. 
In our work, the models are trained to predict words and documents in a sequence with maximum likelihood. We optimize the models using stochastic gradient learning, a flexible and powerful optimization framework suitable for the considered large-scale problems in an online setting where new samples arrive sequentially. 

Vector representations of documents and words learned by our model are useful for various applications of critical importance to online businesses. For example, by means of simple nearest-neighbor searches in the joint vector space between document and word representations, we can address a number of important tasks: 1) given a query keyword, search for similar keywords to expand the query (useful in the search product); 2) given a keyword, search for relevant documents such as news stories (useful in document retrieval); 3) given a document, retrieve similar or related documents (useful for news stream personalization and document recommendation); and 4) automatically generate related words to tag or summarize a given document (useful in native advertising or document retrieval). All these tasks are essential elements of a number of online applications, including online search, advertising, and personalized recommendation. In addition, as we show in the experimental section, learned document representations can be used to obtain state-of-the-art document classification results. 

The proposed approach represents a step towards automatic organization, semantic analysis, and summarization of documents observed in sequences. We summarize our main contributions below:
\begin{itemize}
\item  We propose hierarchical neural language model which takes advantage of the context of document sequences and the context of each word within a document to learn their  low-dimensional vector representations. The document contexts can act as an empirical prior which helps learn smoothed representations for documents. This is useful for learning representations of short documents with a few words, for which \cite{le2014distributed} tends to learn poor document representations as separately learned document vectors may overfit to a few words within a specific document. Conversely, the proposed model is not dependent on the document content as much.
\item  The proposed approach can capture inherent connections between documents in the data stream. We tailor our models to analyze movie reviews where a user's preferences may be biased towards particular genres, as well as Yahoo News articles for which we collect click-through logs of a large number of users and learn useful news article representations. The experimental results on retrieval and categorization tasks demonstrate effectiveness of the proposed model.
\item  Our learning framework is flexible and it is straightforward to add more layers in order to learn additional representations for related concepts. We propose extensions of the model and discuss how to learn explicit distributed representations of users on top of the basic framework. The extensions can be applied to personalized recommendation and social relationship mining.
\end{itemize}


\section{Related work}
The related work is largely centered on the notion of neural language models \cite{bengio2003neural}, which improve generalization of the classic $n$-gram language models by using continuous variables to represent words in a vector space. This idea of distributed word representations has spurred many successful applications in natural language processing. More recently, the concept of distributed representations has been extended beyond pure language words to a number of  applications, including modeling of phrases \cite{mikolov2013distributed}, sentences and paragraphs \cite{le2014distributed}, relational entities \cite{bordes2013translating,socher2013reasoning}, general text-based attributes \cite{kiros2014multiplicative}, descriptive text of images \cite{kiros2013multimodal}, online search sessions \cite{djuric2014icdm}, smartphone apps \cite{yates2015wsdm}, and nodes in a network \cite{perozzi2014deepwalk}.
 
\subsection{Neural language models}
Neural language models take advantage of a word order, and state the same assumption as $n$-gram models that words closer in a sequence are statistically more dependent. Typically, a neural language model learns the probability distribution of next word given a fixed number of preceding words which act as the context. More formally, given a word sequence $(w_1, w_2, \ldots, w_T)$ in a training data, the objective of the model is to maximize log-likelihood of the sequence,
\begin{equation}
\mathcal{L} = \sum_{t=1}^T \log \P(w_t|w_{t-n+1}:w_{t-1}),
\end{equation}
where $w_t$ is the $t^{\text{\scriptsize th}}$ word in the sequence, and $w_{t-n+1}:w_{t-1}$ represents a sequence of successive preceding words $(w_{t-n+1}, \ldots, w_{t-1})$  that act as the context to the word $w_t$. A typical model architecture to approximate probability distribution $\P(w_t|w_{t-n+1}:w_{t-1})$ is to use a neural network \cite{bengio2003neural}. The neural network is trained by projecting the concatenation of vectors for context words $(w_{t-n+1}, \ldots, w_{t-1})$ into a latent representation with multiple non-linear hidden layers and the output soft-max layer comprising $W$ nodes, where $W$ is a size of the vocabulary, where the network attempts to predict $w_t$ with high probability. However, a neural network of large size is challenging to train, and the word vectors are computationally expensive to learn from large-scale data sets comprising millions of words, which are commonly found in practice. Recent approaches with different versions of log-bilinear models \cite{mnih2007three} or log-linear models \cite{mikolov2013efficient} attempt to modify the model architecture in order to reduce the computational complexity. The use of hierarchical soft-max \cite{morin2005hierarchical} or noise contrastive estimation \cite{mnih2012fast} can also help speed up the training complexity. In the following we review two recent neural language models \cite{mikolov2013efficient,mikolov2013distributed} which directly motivated our work, namely continuous bag-of-words (CBOW) and skip-gram (SG) model.

\begin{figure*}
\centering
{\includegraphics[width=0.53\textwidth]{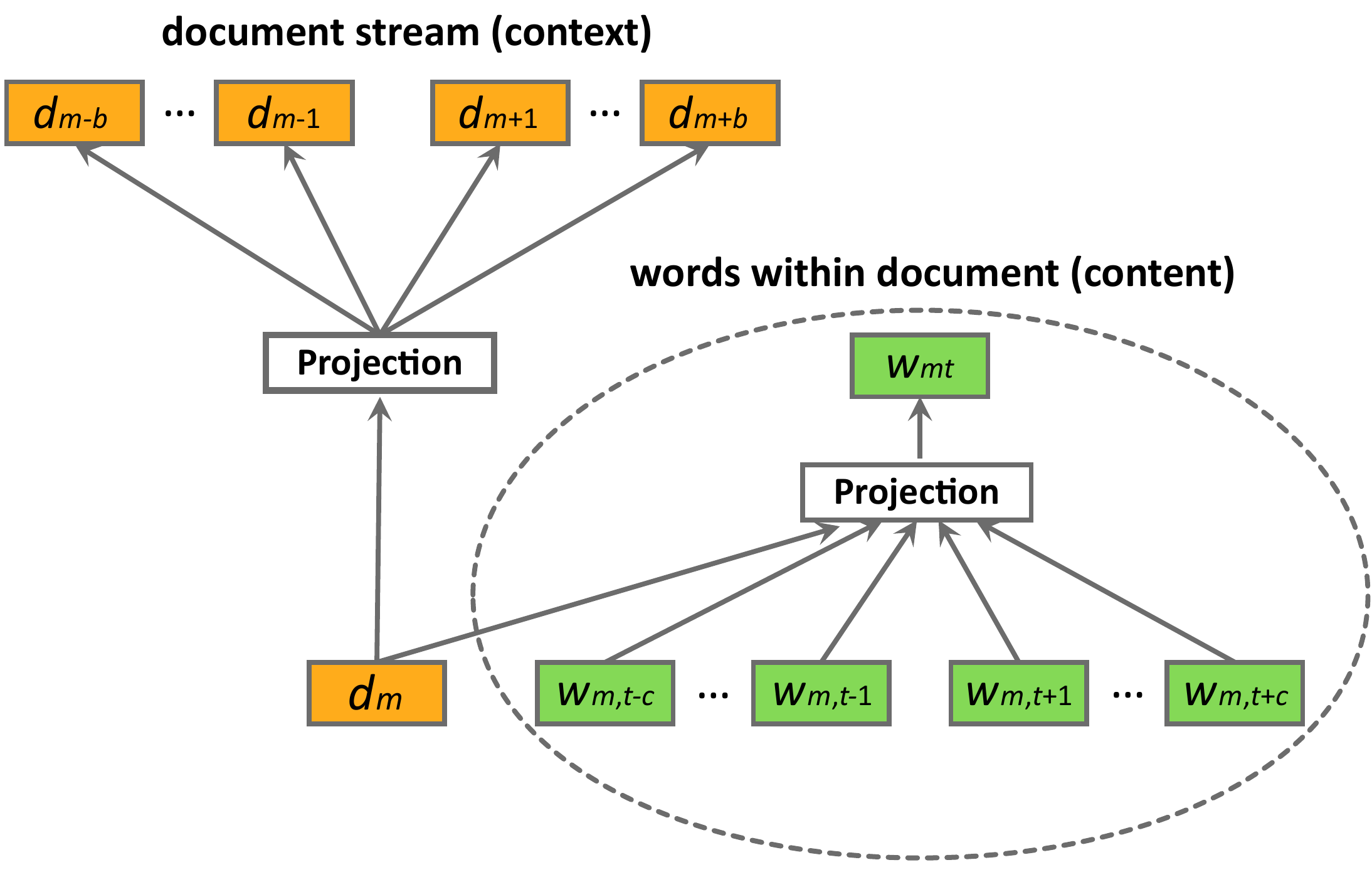}}
\caption{The proposed hierarchical model architecture with two embedded neural languages models ~~~~~~~~~~~~ (orange/left - document vectors; green/right - word vectors)}
\label{fig:architecture}
\end{figure*}

\subsection{Continuous bag-of-words}
The continuous bag-of-words model is a simplified neural language model without any non-linear hidden layers. A log-linear classifier is used to predict a current word based on its preceding and succeeding words, where their representations are averaged as the input. More precisely, the objective of the CBOW model is to maximize the log-likelihood,
\begin{equation}
\mathcal{L} = \sum_{t=1}^T \log \P(w_t|w_{t-c}:w_{t+c}),
\end{equation}
where $c$ is the context length, and $w_{t-c}:w_{t+c}$ is the sub-sequence $(w_{t-c}, \ldots, w_{t+c})$ excluding $w_t$ itself. The probability $\P(w_t|w_{t-c}:w_{t+c})$ is defined using the softmax,
\begin{equation}
\P(w_t |w_{t-c}:w_{t+c}) = \frac{\exp(\bar{\mathbf{v}}^\top\mathbf{v}_{w_t}^\prime)}{\sum_{w=1}^W\exp(\bar{\mathbf{v}}^\top\mathbf{v}_{w}^\prime)},
\end{equation}
where $\mathbf{v}_{w_t}^\prime$ is the output vector representation of $w_t$, and $\bar{\mathbf{v}}$ is averaged vector representation of the context, found as
\begin{equation}
\bar{\mathbf{v}}= \frac{1}{2~ c}\sum_{-c\le j\le c,j\ne 0} \mathbf{v}_{w_{t+j}}.
\end{equation}
Here $\mathbf{v}_{w}$ is an input vector representation of word $w$. 
It is worth noting that CBOW takes only partial advantage of the word order, since any ordering of a set of contextual words will result in the same vector representation.

\subsection{Skip-gram model}
Instead of predicting the current word based on the words before and after it, the skip-gram model tries to predict the surrounding words within a certain distance based on the current one. More formally, skip-gram defines the objective function as the exact counterpart to the continuous bag-of-words model,
\begin{equation}
\mathcal{L} = \sum_{t=1}^T \log \P(w_{t-c}:w_{t+c}|w_t).
\end{equation} 
Furthermore, the model simplifies the probability distribution, introducing an assumption that the contextual words $w_{t-c}:w_{t+c}$ are independent given current word $w_t$,
\begin{equation}
\P(w_{t-c}:w_{t+c}|w_t) = \prod_{-c\le j\le c,j\ne 0} \P(w_{t+j}|w_t),
\end{equation}
with $\P(w_{t+j}|w_t)$ defined as
\begin{equation}
\label{eq:7}
\P(w_{t+j}|w_t) = \frac{\exp(\mathbf{v}_{w_t}^\top \mathbf{v}_{w_{t+j}}^\prime)}{\sum_{w=1}^W \exp(\mathbf{v}_{w_t}^\top \mathbf{v}_{w}^\prime)},
\end{equation}
where $\mathbf{v}_{w}$ and $\mathbf{v}_{w}^\prime$ are the input and output vectors of $w$. Increasing the range of context $c$ would generally improve the quality of learned word vectors, albeit at the expense of higher computation cost. Skip-gram assumes that the surrounding words are equally important, and in this sense the word order is again not fully exploited, similarly to the CBOW model. A wiser strategy to account for this issue would be to sample training word pairs in \eqref{eq:7} less from relatively distant words that appear in the context \cite{mikolov2013efficient}.

\section{Hierarchical language model}
In this section we present our algorithm for joint modeling of streaming documents and the words contained within, where we learn distributed representations for both the documents and the words in a shared, low-dimensional embedding space. The approach is inspired by recently proposed methods for learning vector representations of words which take advantage of a word order observed in a sentence \cite{mikolov2013efficient}. However, and unlike similar work presented by the authors of \cite{le2014distributed}, we also exploit temporal neighborhood of the streaming documents. In particular, we model the probability of observing a particular document by taking into account its temporally-close documents, in addition to conditioning on the content of the document.

\subsection{Model architecture}
In the considered problem setting, we assume the training documents are given in a sequence. For example, if the documents are news articles, a document sequence can be a sequence of news articles sorted in a chronological order in which the user read them. More specifically, we assume that we are given a set $\mathcal{S}$ of $S$ document sequences, with each sequence $s \in \mathcal{S}$ consisting of $N_s$ documents, $s = (d_{1}, d_{2}, \ldots, d_{N_s})$.
Moreover, each document in a stream $d_m$ is a sequence of $T_m$ words, $d_m = (w_{m1}, w_{m2}, \ldots, w_{m,T_m})$. We aim to simultaneously learn low-dimensional representations of streaming documents and language words in a common vector space, and represent each document and word as a continuous feature vector of dimensionality $D$. If we assume that there are $M$ unique documents in the training corpus and $W$ unique words in the vocabulary, then during training we aim to learn $(M+W)\cdot D$ model parameters. 

The context of a document sequence and the natural language context are modeled using the proposed hierarchical neural language model, where document vectors act not only as the units to predict their surrounding documents, but also as the global context of word sequences contained within them. The architecture consists of two embedded layers, shown in Figure~\ref{fig:architecture}. The upper layer learns the temporal context of a document sequence, based on the assumption that temporally closer documents in the document stream are statistically more dependent. We note that temporally does not need to refer to the time of creation of the document, but also to the time the document was read by the user, which is the definition we use in our experiments. The bottom layer models the contextual information of word sequences. Lastly, we connect these two layers by adopting the idea of paragraph vectors \cite{le2014distributed}, and consider each document token as a global context for all words within the document. 

More formally, given set $\mathcal{S}$ of sequences of documents, the objective of the hierarchical model is to maximize log-likelihood of the streaming data,
\begin{equation} \label{maximum}
\begin{aligned}
& \mathcal{L} = \sum_{s \in \mathcal{S}} \Big(\alpha \sum_{d_m \in s} \sum_{w_{mt}\in d_m} \log \P(w_{mt}|w_{m,t-c}:w_{m,t+c}, d_m) +\\
& \sum_{d_m \in s} \big(\alpha \log \P(d_m|w_{m1}:w_{mT}) +  \sum_{-b\le i\le b, i\ne 0} \log \P(d_{m+i}|d_m) \big) \Big), \\
\end{aligned}
\end{equation}
where $\alpha$ is the weight that specifies a trade-off between focusing on minimization of the log-likelihood of document sequence and of the log-likelihood of word sequences (we set $\alpha=1$ in the experiments), $b$ is the length of the training context for document sequences, and $c$ is the length of the training context for word sequences. In this particular architecture, we are using the CBOW model in the lower, sentence layer, and the skip-gram model in the upper, document layer. However, we note that either of the two models can be used in any level of the hierarchy, and a specific choice depends on the modalities of the problem at hand. 

Given the architecture illustrated in Figure \ref{fig:architecture}, probability of observing one of the surrounding documents based on the current document $\P(d_{m+i}|d_m)$ is defined using the soft-max function as given below,
\begin{equation}\label{norm1}
\P(d_{m+i}|d_m) = \frac{\exp(\mathbf{v}_{d_m}^\top \mathbf{v}_{d_{m+i}}^\prime)}{\sum_{d=1}^M \exp(\mathbf{v}_{d_m}^\top \mathbf{v}_{d}^\prime)},
\end{equation}
where $\mathbf{v}_{d}$ and $\mathbf{v}_{d}^\prime$ are the input and output vector representations of document $d$. Furthermore, probability of observing a word depends not only on its surrounding words, but also on a specific document that the word belongs to. More precisely, probability $\P(w_{mt}|w_{m,t-c}:w_{m,t+c}, d_m)$ is defined as
\begin{equation}\label{norm2}
\P(w_{mt} |w_{m,t-c}:w_{m,t+c}, d_m) = \frac{\exp(\bar{\mathbf{v}}^\top\mathbf{v}_{w_{mt}}^\prime)}{\sum_{w=1}^W\exp(\bar{\mathbf{v}}^\top\mathbf{v}_{w}^\prime)},
\end{equation}
where $\mathbf{v}_{w_{mt}}^\prime$ is the output vector representation of $w_{mt}$, and $\bar{\mathbf{v}}$ is the averaged vector representation of the context (including the specific document $d_m$), defined as follows,
\begin{equation}
\label{eq:11}
\bar{\mathbf{v}}= \frac{1}{2~c + 1}(\mathbf{v}_{d_m} + \sum_{-c\le j\le c,j\ne 0} \mathbf{v}_{w_{m,t+j}}).
\end{equation} 
Lastly, we define probability of observing a document given the words contained within $\P(d_m|w_{m1}:w_{m,T_m})$ in a similar way, by reusing equations \eqref{norm2} and \eqref{eq:11} and replacing the appropriate variables.

\subsection{Variants of the model}
In the previous section we presented a typical architecture where we specified language models in each layer of the hierarchy. However, in real-world applications, we could vary the language models for different purposes in a straightforward manner. For example, a news website would be interested in predicting on-the-fly which news article a user would read after a few clicks on some other news stories, in order to personalize the news feed. Then, it could be more reasonable to use directed, feed-forward models which estimate $\P(d_{m}|d_{m-b}:d_{m-1})$, probability of the $m^{\text{\scriptsize th}}$ document in the sequence given its $b$ preceding documents. Or, equivalently, if we want to model which documents were read prior to the currently observed sequence, we can use feed-backward models which estimate $\P(d_{m}|d_{m+1}:d_{m+b})$, probability of the $m^{\text{\scriptsize th}}$ document given its $b$ succeeding documents.

Moreover, it is straightforward to extend the described model for problems that naturally have more than two layers. Let us assume that we have a data set with streaming documents specific to a different set of users (e.g., for each document we also know which user generated or read the document). Then, we may build more complex models to simultaneously learn distributed representations of users by adding additional user layer on top of the document layer. In this setup a user vector could serve as a global context of streaming documents pertaining to that specific user, much like a document vector serves as a global context to words pertaining to that specific document. More specifically, we would predict a document based on the surrounding documents and its content, while also conditioning on a specific user. This variant models $\P(d_{m}|d_{m-b}:d_{m-1}, u)$, where $u$ is an indicator variable for a user. Learning vector representations of users would open doors for further improvement of personalized services, where personalization would reduce to simple nearest-neighbor search in the joint embedding space.

\subsection{Model optimization}
The model is optimized using stochastic gradient ascent, which is very suitable for large-scale problems. However, computation of gradient $\nabla \log \P(w_{mt} |w_{m,t-c}:w_{m,t+c}, d_m)$ in \eqref{maximum} is proportional to the vocabulary size $W$, and of gradients $\nabla \log \P(d_{m+i}|d_m)$ and $\nabla \log \P(d_m|w_{m1}:w_{m,T_m})$ is proportional to the number of training documents $M$. This is computationally expensive in practice, since both $W$ and $M$ could easily reach millions in our applications. 

\begin{table*}
\caption{Movie classification accuracy}
\label{tbl:movie}
\begin{center}
\begin{tabular}{lccccccccc}
{\bf Algorithm}  & {\bf drama} & {\bf comedy} & {\bf thriller} & {\bf romance} & {\bf action} & {\bf crime} & {\bf adventure} & {\bf horror}  \\
\hline \hline
\rowcolor{lightgray}
LDA & 0.5544  & 0.5856  & 0.8158  & 0.8173  & 0.8745  & 0.8685  & 0.8765  & 0.9063\\ 
paragraph2vec  & 0.6367 & 0.6767  & 0.7958  & 0.7919  & 0.8193  & 0.8537  & 0.8524  & 0.8699\\
\rowcolor{lightgray}
word2vec  & 0.7172 & 0.7449  & 0.8102  & 0.8204  & 0.8627  & 0.8692  & 0.8768  & 0.9231\\
HDV & {\bf 0.7274} & {\bf 0.7487}  & {\bf 0.8201}  & {\bf 0.8233}  & {\bf 0.8814}  & {\bf 0.8728}  & {\bf 0.8854}  & {\bf 0.9872}\\
 \bottomrule
\end{tabular}
\end{center}
\end{table*} 

\begin{table*}[t]
\centering
{
  \caption{Nearest word neighbors of selected keywords \label{tb:word-word}}
    \begin{tabular}{c  c c c cccc}
\textbf{boxing} & \textbf{university} &\textbf{movies} &\textbf{batman} &\textbf{woods} & \textbf{hijack} &\textbf{tennis}  &\textbf{messi}  \\
 \hline \hline
 \rowcolor{lightgray}
 welterweight & school     & characters & superman         & tiger   & hijacked      & singles       & neymar     \\
 knockouts & college    & films      & superhero            & masters    & whereabouts   & masters        & ronaldo   \\
 \rowcolor{lightgray}
 fights &professor & studio     & gotham & holes       & transponders  & djokovic     & barca      \\
 middleweight & center     & audiences  & comics         & golf  &  autopilot     & nadal        & iniesta     \\
 \rowcolor{lightgray} 
 ufc & students   & actors     & trilogy         & hole      &  radars        & federer    & libel           \\
heavyweight & national & feature & avenger & pga & hijackers & celebration  & atletico  \\
 \rowcolor{lightgray}
bantamweight & medical & pictures & sci&  classic&  turnback & sharapova & cristiano  \\
greats & american & drama & sequel & par & hijacking & atp  & benzema \\
 \rowcolor{lightgray}
wrestling & institute & comedy & marvel & doral & decompression & slam & argentine  \\
amateur & california & audience & prequel & mcilroy&  baffling & roger  & barcelona  \\

\bottomrule                                                                                                                                         
    \end{tabular}
}
\end{table*}

An efficient alternative that we use is hierarchical soft-max \cite{morin2005hierarchical}, which reduces the time complexity to $\mathcal{O}\big(R\log(W) + bM\log(M)\big)$ in our case, where $R$ is the total number of words in the document sequence. Instead of evaluating every distinct word or document during each gradient step in order to compute the sums in equations \eqref{norm1} and \eqref{norm2}, hierarchical softmax uses two binary trees, one with distinct documents as leaves and the other with distinct words as leaves. 
For each leaf node, there is a unique assigned path from the root which is encoded using binary digits. To construct a tree structure the Huffman encoding is typically used, where more frequent documents (or words) in the data set have shorter codes to speed up the training. The internal tree nodes are represented as real-valued vectors, of the same dimensionality as word and document vectors. More precisely, hierarchical soft-max expresses the probability of observing the current document (or word) in the sequence as a product of probabilities of the binary decisions specified by the Huffman code of the document as
\begin{equation}
\P(d_{m+i}|d_m) = \prod_l \P(h_l|q_l, d_m),
\end{equation}
where $h_l$ is the $l^{\text{\scriptsize th}}$ 
bit in the code with respect to $q_l$, which is the $l^{\text{\scriptsize th}}$ node in the specified tree path of $d_{m+i}$. The probability of each binary decision is defined as follows,
\begin{equation}
\P(h_l = 1|q_l, d_m) =  \sigma (\mathbf{v}_{d_m}^\top \mathbf{v}_{q_{l}}),
\end{equation}
where $\sigma(x)$ is the sigmoid function, and $\mathbf{v}_{q_{l}}$ is the vector representation of node $q_l$. It can be verified that $\sum_{d=1}^M \P(d_{m+i} = d|d_m) = 1$, and hence the property of probability distribution is preserved. We can compute $\P(d_m|w_{m1}:w_{m,T_m})$ in the same manner. Furthermore, we similarly express $\P(w_{mt} |w_{m,t-c}:w_{m,t+c}, d_m)$, but with construction of a separate Huffman tree pertaining to the words.

\section{Experiments}
In this section we describe experimental evaluation of the proposed approach, which we refer to as {\it hierarchical document vector} (HDV) model. First, we validated the learned representations on a public movie ratings data set, where the task was to classify movies into different genres. Then, we used a large-scale data set of user click-through logs on a news stream collected on Yahoo servers to showcase a wide application potential of the HDV algorithm. In all experiments we used cosine distance to measure the closeness of two vectors (either document or word vectors) in the common low-dimensional embedding space. 

\subsection{Movie genre classification}
In the first set of experiments we validated quality of the obtained distributed document representations on a classification task using a publicly available data set. We note that, although we had access to a proprietary data set discussed in Section \ref{sect:exp2} which served as our initial motivation to develop the HDV model, obtaining a similar public data set proved to be much more difficult. 
To this end, we generated such data by combining a public movie ratings data set MovieLens 10M\footnote{grouplens.org/datasets/movielens/, accessed March 2015}, consisting of movie ratings for around $10{,}000$ movies generated by more than $71{,}000$ users, with a movie synopses data set from Internet Movie DataBase (IMDB) that was found online\footnote{ftp.fu-berlin.de/pub/misc/movies/database/, March 2015}. Each movie in the data set is tagged as belonging to one or more genres, such as ``action", ``comedy", or ``horror". Then, following terminology from the earlier sections, we viewed movies as ``documents" and synopses as ``document content". The document streams were obtained by taking for each user the movies that they rated $4$ and above (on the scale from 1 to 5), and ordering them in a sequence according to a timestamp of the rating. The described preprocessing resulted in $69{,}702$ document sequences comprising $8{,}565$ distinct movies, with an average synopsis length of $190.6$ words.

\begin{table*}[ht]
\centering
{\small
  \caption{Most related news stories retrieved for a given keyword}
 \label{tb:word-document}
    \begin{tabular}{ c  c }
    \textbf{movies}    &\textbf{tennis}     \\                                                  
\hline         \hline                                                                                                                   
 \rowcolor{lightgray}
'American Hustle,' 'Wolf of Wall Street' Lead Nominations                      &Tennis-Venus through to third round, Li handed walkover  \\
3 Reasons 'Jurassic World' Is Headed in the Right Direction                      &Nadal rips Hewitt, Serena and Sharapova survive at Miami         \\
 \rowcolor{lightgray}
Irish Film and TV academy lines up stellar guest list for awards                 &Williams battles on at Sony Open in front of empty seats         \\
10 things the Academy Awards won't say     &Serena, Sharapova again on Miami collision course                \\
 \rowcolor{lightgray}
I saw Veronica Mars, thanks to \$35 donation, 2 apps \& \$8 ticket                                         &Wawrinka survives bumpy start to Sony Open                       \\
\bottomrule
\textbf{boxing}       &\textbf{hijack}                                                            \\        
\hline      \hline                     
 \rowcolor{lightgray}
Yushin Okami's Debut for WSOF 9 in Las Vegas                     &Thai radar might have tracked missing plane                                         \\      
UFC Champ Jon Jones Denies Daniel Cormier Title Shot Request                                              &Criminal probe under way in Malaysia plane drama                                    \\      
 \rowcolor{lightgray}
UFC contender Alex Gustafsson staring at a no-win situation                                    &Live: Malaysia asks India to join the expanding search                              \\      
Alvarez back as Molina, Santa Cruz defend boxing titles                                             &Malaysia dramatically expands search for missing jet                                \\      
 \rowcolor{lightgray}
Anthony Birchak Creates MFC Championship Ring                   &Malaysia widening search for missing plane, says minister                           \\      
\bottomrule
\textbf{university}       &\textbf{entertainment}  \\
\hline   \hline 
 \rowcolor{lightgray}
The 20 Public Colleges With The Smartest Students & `American Hustle,' `Wolf of Wall Street' Lead Nominations \\
Spring storm brings blizzard warning for Cape Cod & Madison Square Garden Company Buys 50\% Stake in Tribeca \\  
 \rowcolor{lightgray}
U.S. News Releases 2015 Best Graduate Schools Rankings & Renaissance Technologies sells its Walt Disney position \\
No Friday Night Lights at \$60 Million Texas Stadium & News From the Advertising Industry \\  
 \rowcolor{lightgray}
New Orleans goes all in on charter schools & 10 things the Academy Awards won't say \\
\bottomrule
\end{tabular}
}
\end{table*}

\begin{table*}[t]
\centering
{\small
  \caption{Top related words for news stories \label{tb:document-word}}
    \begin{tabular}{ c c }
{\bf News articles} & {\bf Related keywords}\\
    \hline
 \rowcolor{lightgray}
  &hardcourts biscayne sharapova nadal nishikori \\
\rowcolor{lightgray}
\multirow{-2}{*}{Serena beats Li for seventh Miami crown}& aces unforced walkover angelique threeset \\
\hline
 &isas pensioners savers oft annuity \\
\multirow{-2}{*}{This year's best buy ISAs} &  isa pots taxfree nomakeupselfie allowance\\
\hline
 \rowcolor{lightgray}
 &mwc quadcore snapdragon oneplus ghz \\
 \rowcolor{lightgray}
\multirow{-2}{*}{Galaxy S5 will get off to a painfully slow start in Samsung's home market} &  appleinsider samsung lumia handset android\\
\hline
 &reboot mutants anthology sequels prequel \\
\multirow{-2}{*}{`Star Wars Episode VII': Actors Battle for Lead Role (EXCLUSIVE)}& liv helmer vfx villains terminator\\
\hline
 \rowcolor{lightgray}
 &lesbian primaries rowse beshear legislatures \\
 \rowcolor{lightgray}
\multirow{-2}{*}{Western U.S. Republicans to urge appeals court to back gay marriage}& schuette heterosexual gubernatorial stockman lgbt\\
\hline
&postcold rulers aipac redrawn eilat \\
\multirow{-2}{*}{Russian forces `part-seize' Ukrainian missile units}  &   warheads lawlessness blockaded czechoslovakia ukranian\\
\hline
 \rowcolor{lightgray}
 &catholics pontiff curia dignitaries papacy \\
 \rowcolor{lightgray}
\multirow{-2}{*}{Pope marks anniversary with prayer and a tweet}& xvi halal theology seminary bishops\\
\hline
 &berkowitz moelis erbey gse cios \\
\multirow{-2}{*}{Uncle Sam buying mortgages? Who knew?}& lode ocwen nationstar fairholme subsidizing\\
\hline
 \rowcolor{lightgray}
 &carbs coconut cornstarch bundt vegetarian \\
 \rowcolor{lightgray}
\multirow{-2}{*}{5 Foods for Healthier Skin} & butter tablespoons tsp dieters salad\\
\hline
 &beverley bynum vogel spoelstra shootaround \\
\multirow{-2}{*}{Dwyane Wade on pace to lead all guards in shooting}& dwyane nuggets westbrook bledsoe kobe\\
\bottomrule                                                                                                                                         
    \end{tabular}
}
\end{table*}

Let us discuss several explicit assumptions that we made while generating the movie data set. First, we retained only high-rated movies for each user in order to make the data less noisy, as the assumption is that the users are more likely to enjoy two movies that belonged to the same genre, than two movies coming from two different genres. Thus, by removing low-rated movies we aim to keep only similar movies in a single user's sequence. As we show in the remainder of the section, the experimental results indicate that the assumption holds true. In addition, we used the timestamp of a rating as a proxy for a time when the movie was actually watched. Although this might not always hold in reality, the empirical results suggest that the assumption was reasonable for learning useful movie and word embeddings.

\begin{table*}[ht!]
\centering
{\small
  \caption{Titles of retrieved news articles for given news examples \label{tb:document-document}}
    \begin{tabular}{c }
\textbf{Russian forces `part-seize' Ukrainian missile units}   \\                                
\hline \hline        
 \rowcolor{lightgray}                                                                                                                        
Russia says cannot order Crimean `self-defense' units back to base   \\
Russia says Ukraine ``hallucinating" in warning of nuclear risks      \\
 \rowcolor{lightgray}
Lavrov snubs Ukrainian but says talks will continue                         \\
Kiev must have say in Crimea's future: US                                    \\
 \rowcolor{lightgray}
Crisis in Ukraine: After day of Paris talks, a dramatic change in tone                  \\
\hline                                                                                                                                       
\textbf{Galaxy S5 will get off to a slow start in Samsung's home market}  \\
\hline \hline
 \rowcolor{lightgray}
New specs revealed for one of 2014's most intriguing Android phones                  \\ 
LG G Pro 2 review: the evolutionary process                              \\ 
 \rowcolor{lightgray}
$[$video$]$ HTC wins smartphone of the year                                            \\ 
Samsung apparently still has a major role in Apple's iPhone 6                         \\  
 \rowcolor{lightgray}
Samsung's new launch, the Galaxy S5, lacks innovative features                              \\
\hline 
\textbf{This year's best buy ISAs}\\                                
\hline \hline        
 \rowcolor{lightgray}                                                                                                                        
Savings rates 'could rise' as NS\&I launch new products \\
How to use an Isa to invest in property                                            \\
 \rowcolor{lightgray}
Pensions: now you can have a blank canvas - not an annuity                         \\
Ed Balls' Budget Response Long on Jokes, a Bit Short on Analysis                   \\
 \rowcolor{lightgray}
Half a million borrowers to be repaid interest and charges                         \\
\hline                                                                                                                                       
 \textbf{Western U.S. Republicans to urge appeals court to back gay marriage} \\
\hline \hline
 \rowcolor{lightgray}
 Ky. to use outside counsel in gay-marriage case                              \\ 
 Disputed study's author testifies on gay marriage                            \\ 
 \rowcolor{lightgray}
 Texas' Primary Color Battle Begins                                           \\ 
 Eyes on GOP as Texas holds nation's 1st primary                              \\  
 \rowcolor{lightgray}
 Michigan stumbles in court defending same-sex marriage ban                   \\
\bottomrule
   \end{tabular}
}
\end{table*}

We learned movie vector representations for the described data set using the following methods: 1) LDA \cite{blei2003latent}, which learns low-dimensional representations of documents (i.e., movies) as a topic distribution over their synopses; 2) paragraph vector (paragraph2vec) \cite{le2014distributed}, where an entire movie synopsis is taken as a single paragraph; 3) word2vec \cite{mikolov2013distributed}, where movie sequences are used as ``sentences" and movies are used as ``words"; and 4) the proposed HDV method. We used publicly available implementations of online LDA training \cite{hoffman2010online}\footnote{hunch.net/$\sim$vw/, accessed March 2015} and word2vec\footnote{code.google.com/p/word2vec, accessed March 2015}, and used our implementations of paragraph2vec and HDV algorithms. Note that LDA and paragraph2vec take into account only the content of the documents (i.e., movie synopses), word2vec only considers the movie sequences and does not consider the movie synopses and contained natural language in any way, while HDV combines the two approaches and jointly considers and models both the movie sequences and the content of movie synopses.

Dimensionality of the embedding space was set to $D=100$ for all low-dimensional embedding methods (in the case of LDA this amounts to using $100$ topics). The neighborhood of the neural language methods was set to $5$ for both document and word sequences, and the models were trained for $5$ iterations. Once we obtained the document vectors using the abovementioned methods, we used linear Support Vector Machine (SVM) \cite{CC01a} to predict a movie genre. Note that we chose a linear classifier, instead of some more powerful non-linear one, in order to reduce the effect of variance of non-linear methods on the classification performance and help with the interpretation of the results.

The classification results after 5-fold cross-validation are shown in Table \ref{tbl:movie}, where we report results on eight binary one-vs-rest classification tasks for eight most frequent movie genres in the data set. We can see that the neural language models on average obtained higher accuracy than LDA, although LDA achieved very competitive results on the last six tasks. It is interesting to observe that the word2vec algorithm obtained higher accuracy than paragraph2vec despite the fact that word2vec only considered sequences in which the movies were seen without using the movie synopses, and that, unlike word2vec, the paragraph2vec model was specifically designed for document representation. This result indicates that the users have strong genre preferences that exist in the movie sequences which was utilized by word2vec, validating our assumption discussed earlier. Furthermore, we see that the proposed approach achieved higher accuracy than the competing state-of-the-art methods, obtaining on average $5.62\%$ better performance over the paragraph2vec and $1.52\%$ over the word2vec model. This can be explained by the fact that the HDV method successfully exploited both the document content and the relationships in a stream between them, resulting in improved performance.

\subsection{Large-scale document representation}
\label{sect:exp2}
In this section we evaluate the proposed algorithm a large-scale data set collected on Yahoo servers. The data consists of nearly $200{,}000$ distinct news stories, viewed by a subset of company's users from March to June, 2014. 
After pre-processing where the stopwords were removed, we trained the HDV model on more than $80$ million document sequences generated by users, containing a total of $100$ million words with a vocabulary size of $161{,}000$. Considering that the used data set is proprietary and that numerical results may carry business-sensitive information, we first illustrate a wide potential of the proposed method for numerous online applications using qualitative experiments, followed by the document classification task where we show relative performance improvements over the baseline method.

\subsubsection{Keyword suggestion}
\label{sect:key2key}
An important task in many online applications is finding similar or related words given an input word as a query. This is useful in the setting of, for example, search retargeting \cite{grbovic2014generating}, where advertisers bid on search keywords related to their product or service and may use the proposed model to expand the list of targeted keywords with additional relevant keywords. Table \ref{tb:word-word} shows example keywords from the vocabulary, together with their nearest word neighbors in the embedding space. Clearly, meaningful semantic relationships and associations can be observed within the closest distance of the input keywords. For example, for the query word ``batman", the model found that other superheroes such as ``superman" and ``avengers" are related, and also found keywords related to comics in general, such as ``comics", ``marvel", or ``sequel".

\subsubsection{Document retrieval}
Given a query word, one may be interested in finding the most relevant documents, which is a typical task of an online search engine or news webservice perform. We consider the keywords used in the previous section, and find the titles of the closest document vectors in the common embedding space. As can be seen in Table \ref{tb:word-document}, the retrieved documents are semantically related to the input keyword. Interestingly, in some cases it might seem that the document is irrelevant, as, for example, in the case of keyword ``university" and headlines ``Spring storm brings blizzard warning for Cape Cod" and ``No Friday Night Lights at \$60 Million Texas Stadium". However, after closer inspection and a search for the headlines in a popular search engine, we can see that the snow storm from the first headline affected school operations and the article includes a comment by an affected student. Similar search also confirmed that the second article discussed school facilities and an education fund. Although the titles may be misleading, we can see that the both articles are of interest to users interested in the keyword ``university", as our model correctly learned from the actual user sessions.

We note that the proposed approach differs from the traditional information retrieval methods due to the fact that retrieved documents do not necessarily need to contain the query word, as examplified in Table \ref{tb:word-document} in a case of the keyword ``boxing". As we can see, the HDV method found that the articles discussing UFC (Ultimate Fighting Championship) and WSOF (World Series of Fighting) events are related to the sport, despite the fact that neither of them specifically mentions the word ``boxing". 

\subsubsection{Document tag recommendation}
An important task in online news services is automatic document tagging, where, given a news article, we assign a list of relevant keywords (called {\it tags}) that explain the article content. Tagging is very useful in improving the document retrieval systems, document summarization, document recommendation, contextual advertising (tags can be used to match display ads shown alongside the article), and many other applications. Our model is suitable for such tasks due to the fact that document and word vectors reside in the same feature space, which allows us to reduce complex task of document tagging to a trivial $K$-nearest-neighbor search in the joint embedding space.

Using the trained model, we retrieved the nearest words given a news story as an input. In Table \ref{tb:document-word} we give the results, where titles of the example news stories are shown together with their lists of nearest words. We can see that the retrieved keywords often summarize and further explain the documents. For example, for the news article ``This year's best buy ISAs", related to Individual Savings Account (ISA), the keywords include ``pensioners" and ``taxfree", while in the mortgage-related example (``Uncle Sam buying mortgages? Who knew?") keywords include several financial companies and advisors (e.g., Nationstar, Moelis, Berkowitz).

\subsubsection{Document recommendation}
In the document recommendation task, we search for the nearest news articles given a news story. The retrieved articles can be provided as a reading recommendations for users viewing the query news story. We give the examples in Table~\ref{tb:document-document}, where we can see that relevant and semantically related documents are located nearby in the latent vector space. For example, we can see that the nearest neighbors for article related to the 2014 Ukraine crisis are other news stories discussing the political situation in Ukraine, while for the article focusing on Samsung Galaxy S5 all nearest documents are related to the smartphone industry. 

\subsubsection{News topic classification}
Lastly, we used the learned representations to label news documents with $19$ first-level topic tags from the company's internal interest taxonomy (e.g., taxonomy includes ``home \& garden" and ``science" tags). We used linear SVM to predict each topic separately, and the average improvement over LDA after 5-fold cross-validation is given in Table \ref{tbl:news_doc}. We see that the proposed method outperformed the competition on this large-scale problem, strongly confirming the benefits of HDV for representation of streaming documents.

\section{Conclusion}
We described a general unsupervised learning framework to model the latent structure of streaming documents,
that learns low-dimensional vectors to represent documents and words in the same latent space. Our model exploits the
context of documents in streams and learns representations that can capture temporal co-occurrences of documents and statistical patterns of the words contained within. The method was verified on a movie classification task, outperforming the current state-of-the-art by a large margin. Moreover, experiments on a  large-scale data set comprising click-through logs on Yahoo News demonstrated effectiveness and wide applicability of the proposed neural language approach.

\begin{table}[t!]
\caption{Relative improvement over LDA}
\label{tbl:news_doc}
\begin{center}
\begin{tabular}{lc}
\rule{0pt}{2.5ex}{\bf Algorithm} &  {\bf Avg. accuracy improvement}  \\
\hline \hline
\rowcolor{lightgray}
LDA & 0.00\% \\ 
paragraph2vec  & 0.27\% \\
\rowcolor{lightgray}
word2vec  & 2.26\%  \\
HDV & {\bf 4.39\%} \\
 \bottomrule
\end{tabular}
\end{center}
\end{table}

\balance
%
\bibliographystyle{abbrv}
{
\bibliography{frp0984-djuric-doc2vec}  
}

\end{document}